\documentclass[letterpaper]{article}

\usepackage[numbers]{natbib}
\usepackage{alifeconf}  
\usepackage{amsmath, amsthm}
\usepackage{url,hyperref,cleveref}
\usepackage{booktabs}
\usepackage{float}

\usepackage[most]{tcolorbox}

\usepackage{graphicx} 
\usepackage{xcolor}
\usepackage{etoolbox}
\usepackage{comment}

\newtheorem{definition}{Definition}

\newtoggle{final}
\togglefalse{final} 

\title{From Text to Life: On the Reciprocal Relationship between Artificial Life and Large Language Models}

\author{
    Eleni Nisioti$^{\dag, 1}$,
    Claire Glanois$^{\dag, 1}$, 
    Elias Najarro$^1$, 
    Andrew Dai$^2$, 
    Elliot Meyerson$^3$, \\
   {\Large  
    Joachim Winther Pedersen$^1$, 
    Laetitia Teodorescu$^4$, 
    Conor F. Hayes$^5$, 
    Shyam Sudhakaran $^1$,
   Sebastian Risi $^1$}  \\
    \mbox{}\\
    $^\dag$ Equal Contribution
    $^1$ IT University, Denmark 
    $^2$ Aleph Alpha @ IPAI, Germany \\
    $^3$ Cognizant AI Labs, San Francisco, USA
    $^4$  Inria 
    $^5$ Lawrence Livermore National Lab, USA\\
    \{enis, clgl, sebr\}@itu.dk
} 

\date{February 2024}

\begin{document}

\maketitle

\begin{abstract}
Large Language Models (LLMs) have taken the field of AI by storm, but their adoption in the field of Artificial Life (ALife) has been, so far, relatively reserved. In this work we investigate the potential synergies between LLMs and ALife, drawing on a large body of research in the two fields.
We explore the potential of LLMs as tools for ALife research, for example, as operators for evolutionary computation or the generation of open-ended environments. Reciprocally, principles of ALife, such as self-organization, collective intelligence and evolvability can provide an opportunity for shaping the development and functionalities of LLMs, leading to more adaptive and responsive models. By investigating this dynamic interplay, the paper aims to inspire innovative crossover approaches for both ALife and LLM research. Along the way, we examine the extent to which LLMs appear to increasingly exhibit properties such as emergence or collective intelligence, expanding beyond their original goal of generating text, and potentially redefining our perception of lifelike intelligence in artificial systems.


\end{abstract}
%


\vspace{-2mm}
\section{Introduction}
\begin{figure}[t]
\centering
\includegraphics[width=0.9\columnwidth]{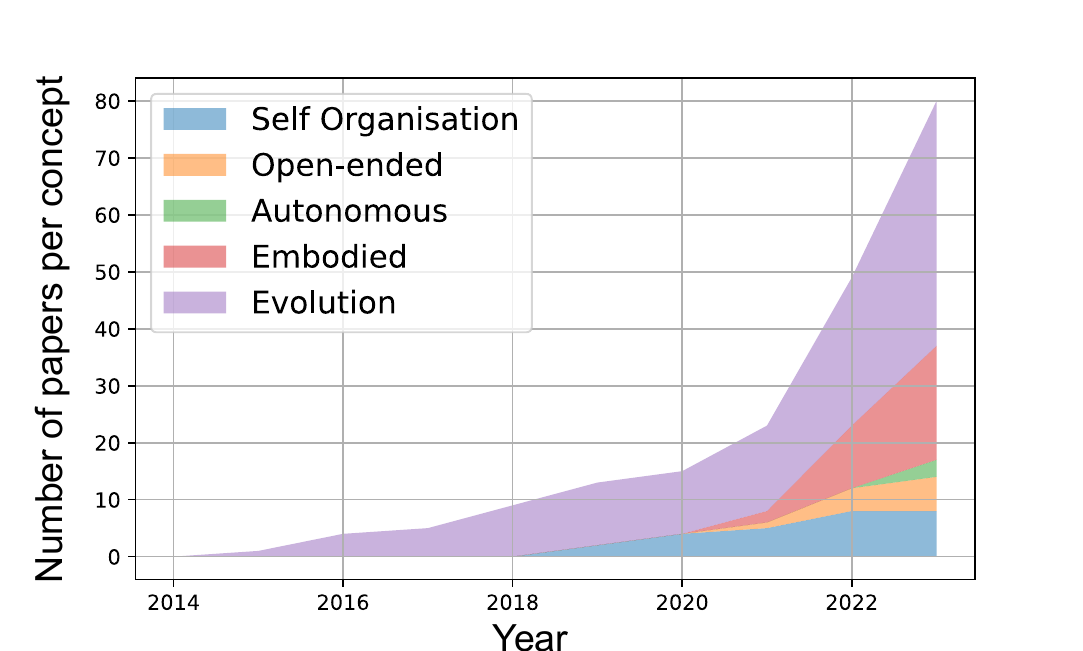}
\caption{Number of papers published at \textit{NeurIPS} that incorporate different lifelike properties.}
\label{fig:neurips_vs_alife}
\end{figure}

Artificial life (ALife) refers to the study and creation of lifelike systems using computer models, algorithms, and, sometimes, physical hardware. Its goal is to understand the fundamental principles of living systems, to explore the processes underlying life, and to develop new forms of life that may exhibit lifelike behaviours. ALife research typically involves simulating and studying various aspects of biological life, such as evolution, self-organization, adaptation, reproduction, and emergence. These simulations may range from simple models of individual organisms to complex ecosystems with multiple interacting species~\cite{dorin2024artificial, kim2006comprehensive, aguilar_past_2014}.  Recently, the Artificial Intelligence (AI) community has started to increasingly embrace such ALife concepts  (see Fig.~\ref{fig:neurips_vs_alife}).


Much of the current focus in AI research has shifted to Large language models (LLMs).
Beyond their impressive ability to generate text in natural language, LLMs are part of an emerging debate on their ability to act as agents that emulate aspects of human behaviour.
Computationally, both ALife forms and LLMs can be seen as autoregressive models capable of processing information and producing complex sequences of patterns. 
However, modern neural network architectures such as the Transformer~\cite{vaswani_attention_2023}—the architecture underpinning LLMs— have been shown to have sufficient computational expressivity to, not only model complex sequences, but also to serve as general-purpose computers~\citep{Giannou2023Jan, Li2023Jul}. This parallelism between the information processing mechanisms of artificial life-forms and transformers, particularly their concurrent processing capabilities, has been underexplored in the ALife literature~\footnote{Last year's ALife conference featured only one LLM paper.}.

\vspace{-1mm}

\begin{figure*}
\centering
\includegraphics[width=0.9\textwidth]{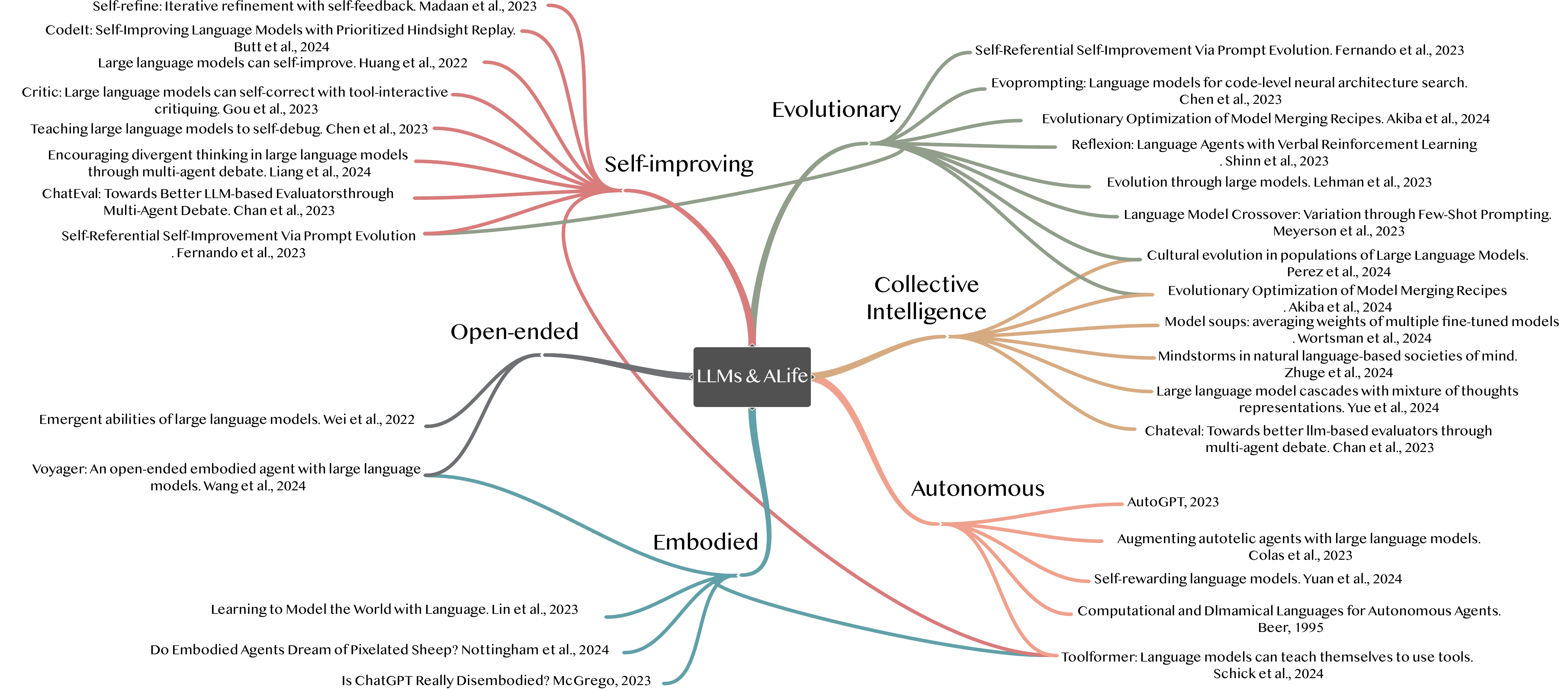}
\caption{Selection of papers on LLMs where some ALife-related concepts are incorporated as key components.}
\label{fig:splash}
\end{figure*}

This paper takes a deeper look into the current —and potential— interplay between these two fields of research, investigating the mutual benefits and potential advancements arising from their intersection.
First, we aim to uncover how LLMs can be harnessed as potent tools within ALife research, for example by enacting powerful new mutation operators for evolutionary computation. Second, we believe that the principles of ALife can reciprocally illuminate and help to expand the functionalities of LLMs. 
Finally, the recent emergence of agents powered by LLMs ---with diverse architectures, sensory modalities, or even social structures known as \emph{LLM Agents}--- raises a provocative question: \emph{could LLM agents themselves be considered a form of ALife}? 
While the view of LLMs as \textit{stochastic parrots} is a sensible and important  critique~\cite{Bender2021Mar}, considering LLM agents as a possible evolvable lifeform in the space of morphologies expressed by their genotype (e.g.\ weights, architecture), with a phenotype (e.g.\ behavior), adaptive sensory inputs and motor outputs ~\citep{aghajanyan2023scaling, muennighoff2024generative}, and adaptive social structures, may challenge our assumptions about life-as-it-could-be~\citep{langton1989artificial}.
By viewing LLM agents through an ALife lens, it may be possible to identify missing aspects or mechanisms that could improve them, leading to more effective innovation, resource usage, and emergent dynamics~\cite{belew1991artificial, eldridge2008manipulating, Langton1989ArtificialLife}.
\section{Background}

\subsection{Artificial Life}
While a review of ALife is beyond our scope (cf.~\cite{dorin2024artificial, kim2006comprehensive, aguilar_past_2014}), we aim here to provide relevant background on fundamental questions in this field. 
The study of biological life has been dominated by two distinct perspectives: (\textbf{a}) The \emph{evolutionary perspective}, exemplified by Neo-Darwinism, states that a system implements life as long as it exhibits the properties of multiplication, variation, and heredity and can be extended to consider other processes, such as development and symbiogenesis~\citep{john_artificial_1999}.
This paradigm focuses on the evolution of open-endedness and/or complexity, given basic evolutionary operators~\citep{MaynardSmith1986-SMITPO-65,bedau_classification_1998}. 
Examples include systems of self-replicating, evolving individuals such as Terraria~\citep{ray_approach_1991} and Avida~\citep{ofria_avida_2004}.
(\textbf{b}) The \emph{ecological perspective}, exemplified by the autopoietic systems popularized by Varela~\citep{varela_principles_1979,varela_autopoiesis_1974}, focusing on the ability of individuals, viewed as structures engaged in energy and matter exchange with their environment, to maintain themselves in the face of environmental perturbations.
Examples systems here include Cellular Automata~\citep{chou_emergence_1997} and Artificial Chemistries~\citep{fontana_arrival_1994}, which examine how self-replication mechanisms emerge from interactions among non-living components.
\emph{What is the objective of ALife as a scientific discipline?} The answer is two-fold:
(\textbf{a}) understanding the essentials of life by simulating bottom-up processes that imitate the biological ones, i.e.\ ALife as the study of life-as-it-is. 
(\textbf{b}): building systems that have lifelike properties and can be viewed as life, independently of their substrate, i.e.\ ALife as the study of \emph{life-as-it-could-be}~\cite{langton1984self,pattee_simulations_1987,Ray1992EvolutionE}.

\vspace{-2mm}
\subsection{Large Language Models}
\vspace{-1mm}
 Language models (LMs) can be understood as probability distributions over sequences of text elements~\citep{wang2020neural,bengio2000neural}. Commonly, LMs aim to predict the next text element, called a token, conditioned on the previous ones~\citep{brown2020language}. This training approach yields generative models of language: tokens are sampled one after the other —i.e.\ \emph{autoregressively}— based on the learned distribution. 
The tipping point in Language Modelling research occurred with the arrival of the massively parallel Transformer architecture\citep{vaswani_attention_2023}\footnote{Replacing notably the slow sequential RNNs and LSTMs~\citep{hochreiter1997long}.}, whose self-attention mechanism allowed both larger-scale training and long-range dependency modelling, and offered compelling performance. 
%
The recent discovery of empirical scaling laws~\citep{kaplan2020scaling, hoffmann2022training} --predicting test perplexity~\citep{jelinek1977perplexity} as a power law of a model’s and dataset's size, and correlating perplexity with downstream task performance--, spurred researchers to train large LMs (LLMs)\footnote{Large-scale models for other modalities such as images~\citep{rombach2022high, ramesh2021zero} or video~\citep{bruce_genie_2024}, as well as multimodal models~\citep{radford2021learning} ensued.} to the 100s of billion parameters and to the trillion tokens~\citep{brown2020language,rajbhandari2020zero}, requiring massive compute and internet-scale data. Some emergent capabilities have been discussed~\citep{wei2022emergent,schaeffer2024emergent}, reminiscent of complex systems ~\cite{west2017scale}, such as zero- or few-shot learning ~\cite{brown2020language},\footnote{I.e.when a model swiftly adapts to a novel task description, with only a few examples, without additional training.} 

Steering LMs towards human-preferred responses solely through next-token prediction proved challenging, so additional fine-tuning approaches, such as instruct-LMs that employ instruction-following datasets~\citep{ouyang2022training} and reinforcement learning through human feedback~\citep{christiano2017deep}, were devised.
These fine-tuned models (e.g.\ \emph{ChatGPT}), enjoyed widespread success for general as well as specialized applications~\cite{kaddour2023challenges}. Their versatility led recently to the emergence of LLM agents: agents whose core policy is an LLM that takes actions based on some observations to achieve certain objectives~\citep{wang_survey_2024}. LLM agents can make use of tools and APIs~\citep{schick2024toolformer} providing them 
with new affordances which, combined with some demonstrated LLMs' abilities such as\ planning~\citep{wei2022chain, zelikman2022star, shinn2023reflexion, yao2024tree}, goal-generation~\citep{colas2023augmenting, wang2023voyager}, and long term semantic and procedural memory retrieval~\citep{wang2023voyager}, can result in more complex behaviour.
\section{LLMS AS TOOLS FOR ALIFE}\label{sec:Tools}
LLMs can find diverse applications that stretch far beyond their obvious use in text generation.
Here we review existing works that incorporate LLMs in ALife studies, organizing them into five main threads of research.

\subsection{Artificial Evolution and LLMs}
Artificial evolution is a powerful optimization algorithm for exploring arbitrary search spaces but comes with the downside of slow exploration due to random mutations and uninformed cross-over.
This has so far limited the application domains of techniques such as Genetic Programming~\citep{koza_genetic_1994,lehman_evolution_2022}. 
Here we highlight different ways in which the ALife community incorporated LLMs in evolution.

Evolution through Large Models (ELM) employed an LLM as an intelligent mutation operator for evolving programs that control robotic morphologies \cite{lehman_evolution_2022}.
The LLM embodies the ability of humans to intentionally modify programs to achieve a desired functionality.
This is achieved by fine-tuning the LLM on data available in online code repositories, where humans employ version control to log code modifications, accompanying each modification with an explanation in natural language.
By replacing random mutations with intelligent ones, ELMs can explore robotic morphologies much quicker than genetic programming and can zero-shot produce functional morphologies for a user-defined terrain. 
Another example is the LMX approach~\cite{meyerson_language_2023}, which employs LLMs as intelligent, domain-independent cross-over operators by leveraging their in-context learning abilities. 
The LLM is prompted with examples of genotypes and, due to its pattern-completion ability, acts as a probabilistic model for generating offspring, akin to models employed in classical evolutionary strategies, such as CMA-ES~\cite{hansen_cma_2005}.
EvoPrompting~\citep{chen_evoprompting_2023} and PromptBreeder~\citep{fernando_promptbreeder_2023} employ LLMs as both mutation and cross-over operators and showcase that evolving the prompts for in-context learning can further improve performance.
In a more holistic approach, EvoLLM models the LLM as a black box that embodies an end-to-end evolutionary algorithm~\citep{lange_large_2024}. The LLM is prompted with a set of solutions and their respective fitnesses and is tasked with providing new solutions. 
QDAIF~\citep{bradley_quality-diversity_2023,pourcel2023aces,samvelyan2024rainbow} utilizes LLMs to both vary/rewrite texts and evaluate qualitative attributes of quality/diversity in subjective writing and code, to select for a population of text solutions to be more diverse and refined.

Overall, this line of work has shown that LLMs hold great potential in improving the speed and applicability of evolutionary search~\cite{wu_evolutionary_2024}.
By operating in the space of code, such techniques enjoy the generality of Genetic Programming and have already been employed for optimizing robotic morphologies~\cite{lehman_evolution_2022,meyerson_language_2023}, neural architectures~\cite{nasir_llmatic_2024} and control policies~\cite{meyerson_language_2023}.
Moreover, benefits are reciprocal~\cite{wu_evolutionary_2024}: by subjecting LLM components to evolution~\citep{chen_evoprompting_2023,fernando_promptbreeder_2023} and Quality-Diversity optimization~\cite{lehman_evolution_2022,meyerson_language_2023,nasir_llmatic_2024}, these systems can self-improve, hinting at an open-ended process.  

\subsection{Environment generation through LLMs}
Environments play a central role in open-ended evolution, as they set the limit for the phenotypic complexity a system can exhibit~\citep{soros_necessary_nodate,wang_poet_2019}.
Generating useful environments for studying ALife is as challenging a problem as creating agents that solve them~\citep{clune_ai-gas_2020}. Techniques such as Procedural Content Generation~\citep{shaker_procedural_2016} can automate this process, but face a long-standing challenge; balancing the diversity and originality of environments with controllability.

Recent works in ALife started exploring the automated generation of environments through LLMs~\cite{kumaran_scenecraft_2023}, by encoding pixels as text.
~\citet{todd_level_2023} explored the potential of LLMs in PCG by fine-tuning an LLM on a two-dimensional puzzle game and showcased that, despite lacking biases for spatial arrangements that previous PCG models such as CNNs and CA exploited, LLMs can create diverse and playable games levels ~\citet{sudhakaran_mariogpt_2023} introduced MarioGPT for generating diverse and controllable levels for Super Mario by employing Novelty Search and conditioning the LLM's output on a textual description of the level. \citet{zala_envgen_2024} incorporated LLMS into the curriculum learning paradigm, where the performance of agents is fed back to the environment generation process to avoid its halt.
Generative Interactive Environments (Genie) is a training paradigm that extended the paradigm of foundational models to the generation of Platformer games~\citep{bruce_genie_2024}.
Genie trained a Transformer on a large corpus of virtual worlds described through text, images, and sketches and showcased the potential of this line of research for the large-scale generation of environments.
Code-generating LLMs~\cite{lehman_evolution_2022} can be seen as universal environment generators and, when embedded within an optimization process that encourages diversity~\cite{lehman_evolution_2022} and self-improvement~\cite{fernando_promptbreeder_2023}, can hold great potential for the generation of open-ended environments~\cite{lehman_evolution_2022}.

\subsection{Exploration through LLMs} \label{sec:exploreLLM}
Play or intrinsic motivation is a vital component of exploration in open-ended spaces, facilitating, among others, the discovery of emergent patterns in self-organized systems~\cite{reinke_intrinsically_2020,hamon_discovering_2024,hamann2014evolution,der_playful_2012} and skill-acquisition with RL agents~\cite{forestier2022intrinsically,pere2018unsupervised}.
Similarly to evolution, intrinsically motivated exploration faces the challenge of quickly discovering interesting information in large search spaces.
Here, we highlight works leveraging LLMs for guiding exploration.

~\citet{lin_learning_2023} recommend interpreting the auto-regressive ability of LLMs as a powerful form of self-prediction that can be useful for predicting next states in embodied environments.
LLMs have been found particularly useful for improving the exploration abilities of agents in open-ended tasks.
For example, they can help assess the achievability of goals based on the agent's abilities, maintain libraries of skills, as well as plan by generating goals ~\citep{du_guiding_2023,colas2023augmenting}  and decomposing tasks into achievable sub-goals~\citep{nottingham_embodied_2023, wang_describe_2023,wang2023voyager},


\subsection{LLMs as models of human behaviour}
ALife has long been concerned with understanding and replicating the emergence of collective and individual phenomena in human populations.
Studies are, however, often limited in their complexity and controllability.
At the same time, human notions such as interestingness, surprise, and creativity are elusive to define in a computational way that will enable optimizing ALife systems.
Here, we highlight works showcasing that LLMs have captured biases that extend beyond language to behaviours and notions related to social interactions, economic decision-making, innovation and interestingness.

When prompted to impersonate different characters, a group of LLM agents in a game inspired by The Sims, emerged convincing social interactions~\citep{park2023generative}.
When evaluated for their decision-making in economic studies, LLMs were found to exhibit human-like biases such as fairness and status quo~\cite{acerbi_large_2023}.
LLMs may also hold promise as tools for cultural evolution studies as they exhibit human-like biases in information transmission~\citep{acerbi_large_2023,perez_cultural_2024}.
By capturing the human notion of interestingness, LLMs can be leveraged in open-endedness research~\cite{zhang2023omni}.

Despite being able to model certain aspects of human behaviour, LLMs' ability to model human cognition and psychology is heavily debated~\cite{lewis_using_2024,mitchell_debate_2023,shiffrin_probing_2023,momennejad_evaluating_2023}. 
Criticism questions their ability to use symbolic and complex abstraction and reasoning~\cite{mitchell_comparing_2023}, utilize cognitive maps for planning~\cite{momennejad_evaluating_2023}, their ``understanding'' of language and if they have a Theory of Mind~\cite{ullman_large_2023}.
Yet, despite being in some ways fundamentally distinct, they provide an opportunity for cognitive science as they allow for the manipulation and observation of input data in ways not possible with human subjects~\citep{wong_word_2023,han_inductive_2023}, and open up interesting questions on whether certain behaviors can emerge purely from textual data.



\subsection{LLMs as scientific collaborators} 

AI holds the promise of changing the way we do science~\cite{Sourati2023Oct, Kitano2021Jun, Messeri2024Mar, BibEntry2024Apr}, including ALife research~\cite{Etcheverry2024Feb}, 
LLMs trained on large scientific corpora putatively capture knowledge about the natural world in the form of language correlations~\cite{Petroni2019Sep}. While we acknowledge the debate on whether these correlations can be considered a strong form of knowledge in an epistemological sense, recent research suggests that LLMs hold potential as scientific collaborators~\cite{Lala2023Dec, Hollmann2023May, Liu2023May, park2023generative, Qu2023Jul, Ziems2024}.

\begin{figure*}[h!]
    \centering
    \includegraphics[width=0.9\textwidth]{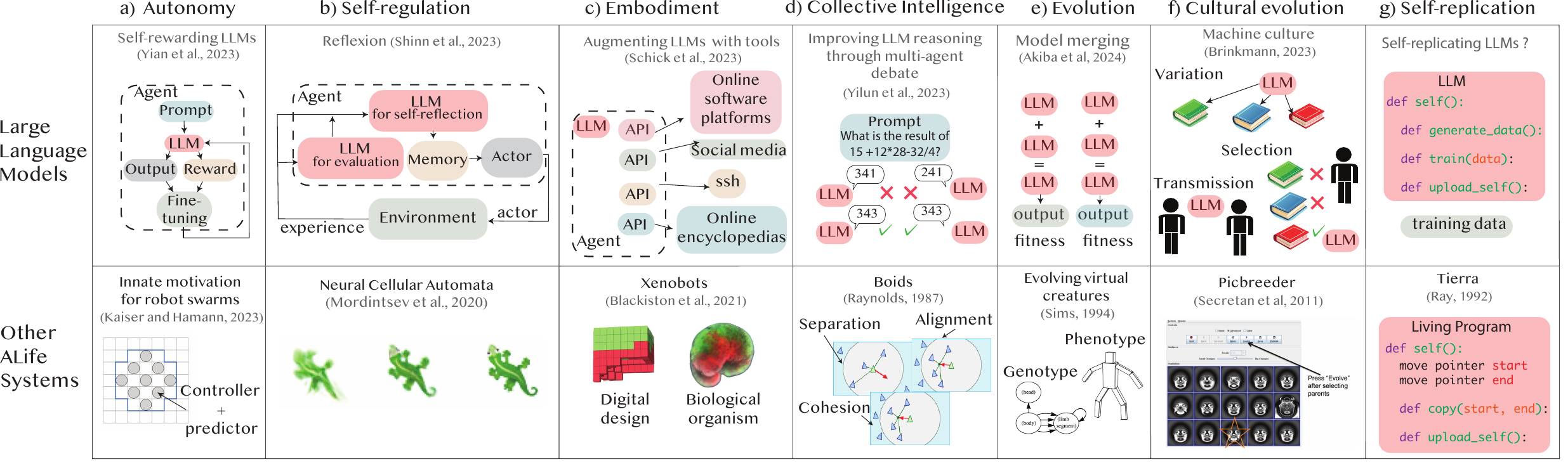}
    \caption{\textbf{Can we view LLMs as ALife systems?} 
    (from left to right) (a) Self-rewarding LLM agents are able to assess their experiences and fine-tune themselves on the best samples~\citep{yuan2024self}. Forms of intrinsic motivation, such as surprise, are employed in robotic swarms~\citep{hamann2014evolution}.
    (b) Under Reflexion, an LLM agent evaluates its own experiences and consolidates them in a long-term memory~\citep{shinn2023reflexion}. In ALife, Neural Cellular Automata act as a model of morphogenesis that displays robustness to perturbations ~\citep{sudhakaran2021growing}. 
    (c) When granted web-access, LLM agents can learn how to use APIs, thus acquiring access to the digital substrate of human society~\citep{schick2024toolformer}. Traditionally, embodiment in ALife refers to the ability of digital organisms, such as Xenobots~\citep{blackiston2021cellular}, to acquire a biological substrate.
    (d) The factuality and reasoning abilities of LLMs seem to improve when they debate with other LLMs ~\citep{du2023improving}.
    In ALife, models such as Boids exemplify the emergence of complex patterns in a collective where individuals follow simple rules~\citep{reynolds1987flocks}.
    (e) Model-merging is a technique for crossing over multiple LLMs into a single one model with improved performance which can be incorporated within an evolutionary loop~\citep{akiba_evolutionary_2024}, a dominant paradigm in ALife~\citep{sims1994evolving} (f) By participating in the generation, transmission and selection of cultural artefacts (e.g.\ texts, images), LLMs and other Generative AI techniques are participating in human cultural evolution~\citep{brinkmann_machine_2023}. Artificial creativity and human-AI collaboration in digital evolution have already engaged ALife~\citep{secretan_picbreeder_2011,lehman_surprising_2020} (g) By viewing an LLM as the composite of its training data and its implementation code, we can envision a future where it replicates, akin to ALife systems such as Tierra~\citep{Ray1992EvolutionE}.}
    \label{fig:llm_as_alife}
\end{figure*}
\section{ALIFE FOR LLMS}\label{ALife4LLM}


Reflecting on the potential of LLMs as instruments for ALife naturally leads us to the reciprocal inquiry: \emph{how can insights from ALife research enhance LLMs?} We review some fundamental characteristics defining (a)life~\cite{aguilar_past_2014, dorin2024artificial, kim2006comprehensive}, and sketch possible parallels with LLM agents~\footnote{We deliberately avoid discussing the metaphysics of these properties, such as emergence or autonomy, assuming they are \emph{epistemological}—not \emph{ontological}—and hence solely exist in the eyes of the observer and not in the systems themselves.} (see Fig.~\ref{fig:splash} for an illustration of these parallels).
\vspace{-2mm}
\paragraph{On Internal States}
\emph{Internal states} of organisms, emerging from the dynamic interactions with their environment, and their broader developmental and evolutionary history, guide their behaviour and decision-making~\cite{boly2008intrinsic, zagha2014neural, miller2022natural}. 
Differently from living organisms, LLMs lack of \emph{self-sustained activity} \cite{lynch1971mechanisms, garrido2007evoked, vernet2020visual,vogels2005signal, buzsaki2006rhythms, santos2022active}\footnote{LLMs remain dormant, waiting for a function call, without cognitive processes running in the background.} 
The internal state of an LLM is entirely a function of the current contents of its context window. A main bottleneck when scaling up Transformer architectures is the fact that the computational requirements of the standard self-attention mechanism  scale quadratically with the length of the context~\cite{tay2022efficientTrans}.
The limited context window  can lead to temporal inconsistencies and erratic behaviour when exceeded due to the discrete capacity\footnote{Architectures like RWKV~\cite{peng2023rwkv} (akin to RNN at inference) seem more appropriate for continual processing than Transformers.}. 
Some efforts in LLM research are currently directed towards addressing these constraints by augmenting them with external memory (e.g.\ with Retrieval-augmented generation)~\cite{gupta2020gmat, lei2020mart, gao2023retrieval}, sometimes including abstractions of the raw observations~\cite{park2023generative, shinn2023reflexion}, or towards more self-sustained activity (via incorporating internal feedback mechanisms, or recurrent LLM streams)~\cite{zhang2023controlling, zhuge2024language}. Further, a wealth of research aims at mitigating the limitations set by the quadratic requirements of the self-attention mechanism~\cite{schuurmans_memory_2023,zaheer2020big, Peng2021RandomFA, peng2023yarn, chen2023extending, liu2023ring, dong2024exploring, xu2024retrieval, jin2024llm}. 

\subsection{Autonomy}
\begin{definition}
\emph{Autonomy} refers to the degree to which a system can `govern' itself (i.e.\ make decisions, take actions) based on its own objectives, internal states, history or processes, without direct external control (e.g.\  intervention, supervision or instruction).  
\end{definition}
While the notion of autonomy is complex and polysemic~\cite{boden2008autonomy, ruiz2004basic, beer1995computational, froese2007autonomy}, in a permissive understanding, most artificial agents and LLM agents~\cite{Yang2023Jun,wang2023jarvis,zhuge2024language,li2024camel,wang2023voyager} can be considered autonomous, from the moment they have some ``control'' over how to act. \emph{Self-sustainability}, i.e.\ requiring the system's ability to manage and regulate the flow of matter and energy, to sustain themselves, 
is often considered out of reach for digital agents. 

Agents that exhibit the ability to \textbf{self-optimize} their behaviors in the absence of external rewards demonstrate a certain degree of autonomy. For instance, \textbf{self-rewarding} agents can adjust their strategies or behaviors by generating their own feedback mechanisms (e.g.\ Fig.~\ref{fig:llm_as_alife}, \cite{kaiser_engineered_2019, hamann2014evolution, yuan2024self}).
Dynamic, self-driven enhancement of LLMs~\cite{pan2023automatically}, leveraging experiences and self-generated data to iteratively refine their model, had broad applications from debugging~\cite{chen2023teaching} to harmlessness/helpfulness~\cite{bai2022constitutional}. \emph{Self-improving} LLM agents, either display simple \emph{behavioral adaptation} via in-context learning or exhibit more \emph{structural adaptation}. The former can gradually refine their outputs~\cite{gou2023critic}
when integrated with an iterative workflow e.g.\ a critique-refine approach~\cite{madaan2024self, chen2023teaching, shinn2023reflexion}.
The latter improve their internal model through self-reinforcing loops, e.g.\ fine-tuning additional parameters on self-generated~\cite{butt2024codeit, gulcehre2023reinforced} or self-assessed~\cite{huang2022large, aksitov2023rest} data, or by optimising their architecture~\cite{zhuge2024language}.
In some cases, training LLMs on such self-generated, synthetic data has been shown to worsen performance, as it may go against prior optimisation or lead to cascading hallucinations~\cite{jiang2024self}. However, combining synthetic with real data can lead to a better cost/performance trade-off~\cite{gerstgrasser_is_2024,llamaopenai}.

\begin{definition}
\emph{Autotelic} systems can learn to represent, generate, select and pursue their own goals, thereby guiding their actions/decisions autonomously.
\end{definition}
Such behaviour ranges from being instinctual, and short-sighted, as in bacterial chemotaxis toward nutrients, ants foraging for food based on pheromone trails, or birds migrating seasonally following instinctual patterns, to being sophisticated and deliberate, as seen in humans contemplating long-term future pathways.
Drawing on essential concepts like \emph{intrinsic motivation} from autonomous developmental learning~\cite{oudeyer_what_2007}, both AI and ALife researchers have proposed agents that could self-generate or self-represent their own goals~\cite{forestier2022intrinsically, pere2018unsupervised, colas2022autotelic}, without supervision, driven by \emph{curiosity}~\cite{parisi2021interesting}, surprise, or \emph{empowerment}~\cite{choi2021variational}, framed in an information-theoretic sense.
ALife works have also investigated how such capacity may speed up language emergence~\cite{cornudella2015intrinsic}, relate to affordances~\cite{aguilera2019quantifying}, or the emergence of collective behaviors ~\cite{hamann2014evolution}, among other things. As previously discussed, 
several LLM agent frameworks propose planning, goal decomposition, or even goal generation~\cite{du_guiding_2023,colas2023augmenting} by leveraging pre-trained LLMs's world model.
Progressing towards more \emph{autonomous} LLM agents may entail enabling them to dynamically refine their mechanisms for goal representation/selection but also self-generate/self-adapt their control and regulatory mechanisms. 

\vspace{-1mm}
 \subsection{Self-Replication} 
\begin{definition}
\emph{Self-Replication} can manifest in two ways: (a) system-level self-replication is the process by which a system makes a copy of itself or its components (b) pattern-level self-replication is when a system makes copies of certain outputs/behaviors it produces.
\end{definition}
By enabling the transmission of information across generations when coupled with evolution, \emph{Self-Replication} is a pivotal feature of Life.
In ALife, self-replication has been an historical quest~\cite{taylor2020rise, sipper1998fifty, penrose1959self}, from von Neumann's automata~\cite{Neumann:66}, Wolfram's~\cite{wolfram1983statistical} or Langton's cellular automata~\cite{langton1984self}, to Kauffman's work on autocatalytic networks~\cite{hordijk2019history} and recent works such as Lenia~\cite{chan2020lenia}. 
These systems exhibit pattern-level self-replication.
In contrast, artificial systems such as computer viruses exhibit system-level self-replication.


 \emph{Pattern-level self-replication} in LLMs can be seen through the memes that LLMs create and spread across our social networks~\cite{brinkmann_machine_2023} 
LLMs Agents show early forms of system-level self-replication, as they offload parts of their cognitive processes by instantiating ``siblings'' to divide and conquer a given task~\cite{Yang2023Jun, hong2023metagpt, zhuge2024language}.
This emergent ability to self-replicate beyond their intended environments, can be seen as a source of risk and is under scrutiny (e.g. see ASL-3 Commitments in~\cite{anthropic2023responsible}).
\vspace{-1mm}
\subsection{Self-Organisation}
\begin{definition}
\emph{Self-organisation} refers to the spontaneous emergence of global order or coordination in a system from local interactions of its parts. That is, self-organisation can be understood in terms of the presence of an attractor driving the dynamics.
\end{definition}
Self-organisation has been observed in physical, biological, chemical, and social systems, from bird flocks to animal fur patterns formation, and crystal growth, via morphogenesis, where cells in a living body divide and specialize to develop into a complex body plan. While its mechanisms are still being investigated, this ability is foundational to enable systems with \emph{self-regulating, self-repairing, self-optimizing, and self-assembling} capabilities (discussed below), adaptivity and resilience.
Self-organisation is an essential inspiration both for ALife systems~\cite{gershenson2018self}, and beyond~\cite{Hopfield1982Apr}, from the early autopoietic models~\cite{varela_autopoiesis_1974}, snowflake formation~\cite{packard1985lattice} or Conway's Game of Life~\cite{bak1989self} to neural cellular automata~~\cite{mordvintsev2020growing, sudhakaran2021growing, chan2018lenia, nichele2017neat} or information theoretic approaches ~\cite{prokopenko2013guided,polani2013information,williams2006homeostatic}. 
\textbf{Self-assembling}, as a path towards more adaptability, has been studied in ALife or AI systems~\cite{nagpal2002programmable, sahin2002swarm, najarro2023towards, pathak2019learning}.

Regarding LLM agents, although recent research hints in that direction investigating locally-interacting agents~\cite{ishibashi2024self}, it does not seem to exhibit self-organised properties. 


\vspace{-3mm}
\paragraph{Emergence}
Both the notion of self-organisation and autonomy is tied to the concept of \emph{emergence}\footnote{The degree of autonomy of agents, whether living or artificial, is correlated to the extent to which their behaviors appear \emph{emergent}.}, which is a central topic both in ALife and biological life. 
In ALife, works have studied the emergence of structure out of chaos, of self-repair properties, of cooperation~\cite{ritz2021sustainable}, of division of labor~\cite{tomko2011many}, of language~\cite{cornudella2015intrinsic}, etc.

While downstream task performance of LLMs has often impressed, the emergent behaviors of LLMs from \emph{scaling} (reminiscent of complex systems~\cite{west2017scale}) are part of a hot debate~\cite{wei2022emergent, schaeffer2024emergent}).
LLMs research significantly diverges from the \emph{bottom-up} approach characteristic of ALife. As of now, many of the properties attributed to LLM Agents in this section, such as self-repair, self-improvement, self-replication, etc. are hard-coded instead of emergent.
\vspace{-3mm}
\paragraph{Self-Regulation}
\begin{definition}
\emph{Self-regulation} refers to the ability of organisms to manage and adjust their internal state and functions in response to internal and external changes.
\end{definition}
Such ability typically involves various regulatory mechanisms (feedback loops), concepts at the core of cybernetics~\cite{beer2002cybernetics}. 
ALife systems, like Xenobots~\cite{blackiston2021cellular} and (neural) cellular automata~\cite{mordvintsev2020growing, sudhakaran2021growing, chan2018lenia}, often display self-regulation properties, such as the ability to  recover from damage. 

Under Reflexion \cite{shinn2023reflexion}, an LLM maintains its own reflective text (from self-assessed experiences) in an episodic memory buffer to induce better decision-making in subsequent trials. Recent studies have also explored the extent to which LLMs may exhibit self-repair \emph{innate} capabilities, specifically examining whether and how components of LLMs compensate upon ablation of certain elements (e.g.\ attention heads~\cite{rushing2024explorations}).
Future research could aim at developing self-repairing LLMs upon damage to some of their components. Advancing the development of additional mechanisms that enhance the \textbf{adaptability} of both the morphologies (architecture) and the code (weights) of LLM Agents in response to internal or external change seems also crucial~\footnote{Some exists (yet with fixed morphologies), as parameter-efficient finetuning~\cite{xu2023parameter}, delta-tuning, or LoRA merging.}.

\subsection{Embedded, Embodied, Enacted, Extended}
\begin{definition}
The 4E cognition framework~\cite{newen2018oxford} posits that cognitive processes in agents are deeply influenced by:
(1) the immediate physical and social environment in which an organism exists (\emph{embedded}), (2) the body's interactions with the world (\emph{embodied}),
(3) the dynamic interactions between an organism and its world (\emph{enacted}), and
(4) can extend beyond the brain to include tools, technologies, and other environmental elements (\emph{extended}).
\end{definition}
 
The 4E framework is widely investigated in ALife~\cite{clark1998being, stepney2007embodiment}. Agents may rely on environmental cues to perform tasks, and learn representations or behaviors which emerge from their actions and interactions with the environment and through sensory-motor couplings.
Occasionally ALife systems may also have a physical body~\cite{blackiston2021cellular, vcejkova2021robots, damiano2023wetware}. 

Regarding \textbf{embeddedness},
LLMs are situated in the sense that their response is highly sensitive to its context, fed as the prompt. 
Although by default very limited, their environment can be -and has been- drastically extended for instance through external and real-time information via Web access and navigation~\cite{gur2023real}, computer access, API extensions~\cite{schick2024toolformer}, etc.
Regarding \textbf{embodiment}, although LLMs or LLM agents often lack a 'tangible body', they may be \emph{physically situated} as pointed out in~\cite{mcgregor2023chatgpt}, which seems the most critical component for 4E. By dynamically focusing on certain parts of its sensory input and adjusting its focus with feedback mechanisms, the attention mechanism in Transformers loosely echoes some form of sensorimotor coupling. To which extent such purely textual sensorimotor contingencies in the case of LLMs do impose certain cognitive distinctions or limitations compared to other forms of cognition remains to be investigated. Tangentially, recent works have grounded LLM agents further in our physical tangible environment, extending some of their sensorial abilities, as multi-modal LLMs~\cite{wu2023next}, API-extended LLM agents~\cite{schick2024toolformer}, or even robotic-bodied LLM agents~\cite{ahn2022i, huang2022inner, brohan2023rt2,yoshida_text_2023}.
LLM Agents can be seen \textbf{enacted} when their cognitive processes and representations still develop through training from their interactions with the environment, as when paired with continual learning (or weakly, with in-context learning). Yet additional plasticity mechanisms could be explored for LLM Agents. 
Lastly, as \textbf{extended} agents, some LLM Agents can outsource some cognitive tasks beyond their own skills, by leveraging existing tools~\cite{bubeck2023sparks, qin2023tool,schick2024toolformer}, or even learning how to create their own tools~\cite{wang2023voyager}. Their tool use often stems from existing knowledge of the world rather than emergent and extrapolative innovation.


\vspace{-1mm}
\subsection{Collective Intelligence}
\begin{definition}
\emph{Collective intelligence} refers to the phenomena of simple local interactions between agents leading to the emergence of complex behaviors.
\end{definition}
The diverse forms of interactions within and between groups of individuals --sometimes resulting in \emph{collective intelligence}-- play a crucial role in the development of both natural and ALife systems. 
Multi-Agent ecosystems have  been highlighted as sources of emergent innovation, within ALife research and beyond~\cite{jaques2019social}, both in collaborative~\cite{panait2005cooperative} and competitive setups~\cite{bansal2017emergent, baker2019emergent, sims1994evolving}. 
Collective Intelligence has raised particular attention in ALife, from flocking agents models~\cite{reynolds1987flocks} to swarm bots~\cite{dorigo2004evolving, rubenstein2014programmable,hamann2018swarm}, also investigating the role of diverse relationships, such as symbiosis~\cite{vostinar2022symbiosis,sunehag2019reinforcement}, parasitism~\cite{hickinbotham2015environmental} and mutualism~\cite{cruse2011egocentric}.

To enhance LLM abilities, many recent works~\cite{li2024camel, liang2023encouraging,chan2023chateval, nair2023dera, zhuge2024language, hong2023metagpt,meta2022human, wang2023jarvis, zeng2022socratic, ishibashi2024self, zhuge2023mindstorms} exploit multi-agents collaborations, for instance multi-agent debate~\cite{liang2023encouraging,chan2023chateval}, even across modalities~\cite{zhuge2023mindstorms}. These works may involve \emph{role specialisations}~\cite{hong2023metagpt,cai2023large} (sometimes dynamically assigned roles~\cite{li2024camel}), diverse  \emph{ecological niches} (e.g.\ different sensorial modalities~\cite{zeng2022socratic}) and diverse \emph{social structures} (sometimes even optimized alongside prompts~\cite{zhuge2024language}). 
However, these multi-LLM-agent frameworks~\footnote{They can also be seen as a step towards nested individual agents, wherein cognition and decision-making occurs across multiple scales, akin to living organisms~\cite{bechtel2021grounding, bravi2015unconventionality}, from cells to society.}, mostly focus on performing short-horizons tasks, with simple in-context social learning (e.g.\ zero/few-shot learning). 
Further exploration could harness enhanced social learning mechanisms, such as reinforcement learning (as in the Economy of Minds envisioned in~\cite{zhuge2023mindstorms}), and adaptive social organization with the aspiration of fostering collective behaviors that are not just marginally superior but qualitatively distinct from both individual behaviors and their aggregation.

\subsection{Evolution}\label{sec:Evolution}
Evolutionary processes, driven by an interplay of \emph{selection}, \emph{variation} (mutation, cross-over), and \emph{transmission} (heredity) processes, across both temporal and spatial scales, have played a pivotal role in the emergence, adaptation, and diversification of life forms, both biological and artificial~\cite{smith1997major}. The importance of \emph{historicity} for behavior, and the interplay of different time scales --ontogenetic, phylogenetic-- should also be noted~\cite{gomez2019life}.
\begin{definition}
\emph{Open-endedness} refers to the capacity of a system to continuously produce novel, and increasingly sophisticated and diverse behaviors, without predefined limit.
\end{definition}
In ALife --and beyond~\cite{team2021open}--, open-endedness is an active area of research~\cite{stanley2019open, packard2019overview, banzhaf2016defining,wang2020enhanced, team2021open,grbic2021evocraft, lehman2008exploiting, lehman2012beyond, soros2014identifying}. Factors like the open-endedness of environment and complexity of phenotypes still hinders progress of weak ALife towards more open-ended systems~\citep{soros_necessary_nodate,bedau_open_2000}.

Akin to genetic operators such as crossover, \emph{model merging} aims to encapsulate the knowledge of multiple models into a single one~\cite{wortsman2022model}, by `merging' them at the parameter level or at the data flow level (e.g.\ transformer blocks). Many approaches utilize a weighted average to combine the model weights into a single model, with hand-tuned operations to find the best model. 
Yet, when combined with evolutionary algorithms~\cite{akiba_evolutionary_2024}, it may enable the automated creation and improvement of LLMs across generations, and has led to state-of-the-art LLMs. 
Analysing the phylogeny of LLMs (how models are combined/fine-tuned to form new models) can be useful towards better understanding them and predicting their performance~\citep{yax2024phylolm}.
The \emph{evolvability} of agents co-evolving with their environment, alongside with open-ended evolution and the interplay of different time scales of adaptive behavior (evolution, lifetime, development, learning) should be further investigated for LLM agents in open-ended environments (e.g.\ real world, web).
 
\paragraph{On LLM Ecologies}

The proliferation of distinct LLMs across dimensions such as size, speciality, and architecture suggests that we are already experiencing a certain LLM ecology~\cite{minaee2024large, zhao2023survey, castano2023exploring}, sometimes referred to as a ``Model Garden''~\footnote{\url{cloud.google.com/model-garden}.}.
`Successful' models, where success denotes economic utility) influence the creation of future generations of models through fine-tuning or design inspiration, at a \emph{pace} which gives us a visceral picture of variation, death, and growth. 
The array of model scales, from the thousands to trillions of parameters, is akin to scales observed in natural ecologies~\cite{white2007relationships}, or ALife ecologies~\cite{charity2023amorphous, earle2023quality, lehman2011evolving}.
Smaller LLMs are currently more plentiful, mirroring natural ecosystems~\cite{woodward2005body}, and have a generally shorter lifespan, akin to the natural ``thermodynamic law''~\cite{speakman2005body}. 

\emph{Compared to other technologies, what makes LLM ecologies particularly suitable for study from an ALife perspective?}
First, LLM agents operate through a common interface --e.g.\ text in, text out--, enabling them to function non-trivially across many environments, akin to the universality of physical embodiment in biology. 
Second, both LLMs and ecosystems are grounded in energy, which is becoming a \emph{critical factor} in their development and use~\cite{castano2023exploring, faiz2024llmcarbon}.
This may lead to the stratification of LLMs across diverse thermodynamic regimes akin to natural and ALife ecosystems~\cite{speakman2005body, earle2023quality, mccormack2007artificial} and task offloading between larger and smaller models~\cite{ding2024hybrid, yue2024large, wang2024survey,cai2023large}.
Third, LLMs are all \emph{made out of the same stuff on which they can operate} (e.g.\ code, Transformer blocks~\cite{zhao2023survey})-- as all organisms are made of DNA, proteins, etc, which grants them the capability to self-improve, self-variate and and self-generate.

\vspace{-1mm}
\subsection{Cultural Evolution}
Both human and non-human entities have the capacity to act and influence their respective ecosystem. LLMs --as other AI~\cite{shin_human_2021, whittaker2021recommender} or technological objects-- can be seen as actors that participate in and co-construct human socio-cultural reality, as advocated by Gilbert Simondon~\cite{simondon2011mode}, and  Actor-Network Theory~\cite{latour2007reassembling} among others. 
This was also witnessed with AI systems, e.g.\ reinforcement learning algorithms integrated within the society effects on human behavior, such as AlphaGo changing the way humans play Go~\cite{shin_human_2021}, or Recommender Systems affecting polarisation and radicalisation~\cite{whittaker2021recommender}.
Increasingly LLMs are mediating our decisions, our ways to consume and share information (e.g.\ as search engines, recommenders,  content generators),  our relationships (e.g.\ as writer assistants), and our very cognitive processes (e.g.\ as collaborators~\cite{vicente2023humans}) --\emph{at a scale rarely matched by other technological objects}. They could have a considerable influence on our behaviors, psychological well-being, skills, 
 social and economic structures, and socio-cultural evolution itself in general~\citep{brinkmann_machine_2023, lehman2023machine, Bender2021Mar, pineiro2023ethical, jakesch2023co}.
\section{Discussion}

In this paper, we highlight the potential reciprocal relationship between ALife and LLMs. On the one hand, our review of works employing LLMs as tools for ALife shows that the emergence of LLMs has opened new avenues for addressing many open questions in ALife~\citep{bedau_open_2000}.
On the other hand,  the methodologies and perspectives from ALife research offer valuable insights for framing the development and functionality of LLMs. The principles of evolvability, self-organisation, and emergence that underpin ALife can inform strategies for designing LLM architectures and training processes, towards models more adaptive in response to shifting environments.

\emph{Are LLMs a form of ALife}?  Some LLM Agents begin to exhibit behaviors akin to artificial life, e.g.\ self-replication, self-repair, collective interaction, tool use, division of labour, planning, goal generation, and being embedded in complex open-ended environments (e.g.\ web). Yet, their capabilities are currently mostly designed top-down instead of being evolvable and emerging, driven by intrinsic motivation, and complex couplings with their ecosystem, while accounting for developmental and evolutionary histories.
Beyond providing a definite answer, the exercise of drawing this analogy may provide new stepping stones for advancing both LLM agents and ALife systems \citep{stanley2018art}.

However, the relationship between ALife and LLMs is not without challenges, including those arising from LLMs alone \cite{kaddour2023challenges}. 
The complexity of LLMs, their dependence on vast and sometimes dubious datasets~\cite{longpre2023data, crawford2021excavating}, and their energy consumption raises questions about their ecological and ethical implications~\cite{Bender2021Mar, pineiro2023ethical}.
The rapid --and thus immature-- profit-driven deployment of LLMs may have a concerning impact on human socio-cultural evolution.
The moral implications of such rapid technological progress have been contemplated by both in AI and ALife.
On one hand, the AI alignment movement is concerned with the controllability and interpretability of the AI ecosystem~\cite{ji_ai_2024}.
On the other hand, ALife researchers ponder the moral responsibility of creating artificial systems that participate in human societies as autonomous agents~\citep{witkowski_ethics_2022,holy-luczaj_hybrids_2021}.
Understanding these implications may benefit from a dialogue between the two communities.

\section*{Acknowlegements}
This project was partially supported by a European Research Council (ERC) grant (GA no. 101045094, project ”GROW-AI”) and 
 a Sapere Aude: DFF Starting Grant (9063-00046B).


\footnotesize
\bibliographystyle{apalike} 
\bibliography{main}

\end{document}